\documentclass[sigconf]{acmart}
\usepackage{multirow}
\usepackage{makecell}

\usepackage{listings}
\usepackage{color}
\usepackage{algorithm}

\AtBeginDocument{%
  }

\acmConference[Arxiv]{}{August}{2023}
\renewcommand\footnotetextcopyrightpermission[1]{}
\usepackage{etoolbox}
\makeatletter
\patchcmd{\authornote}{\g@addto@macro\addresses{\@authornotemark}}{}{}{}
\makeatother

\begin{document}

%%
%% The "title" command has an optional parameter,
%% allowing the author to define a "short title" to be used in page headers.
\title{Relational Contrastive Learning for Scene Text Recognition}

\author{Jinglei Zhang$^{\ast}$, Tiancheng Lin$^{\ast}$, Yi Xu$^{\dagger}$, Kai Chen, Rui Zhang}

\authornote{Both authors contributed equally to this research.}
\authornote{Corresponding author.}
\affiliation{%
  \institution{MoE Key Lab of Artificial Intelligence, AI Institute, Shanghai Jiao Tong University}
  \city{Shanghai}
  \country{China}
}
\email{{zhangjinglei168, ltc19940819, xuyi, kchen, zhang_rui}@sjtu.edu.cn}

% \author{Jinglei Zhang \qquad   Tiancheng Lin  \qquad  YiXu \textsuperscript{\thanks{Corresponding author.}} \qquad  Kai Chen \qquad  Rui Zhang}

% MoE Key Lab of Artificial Intelligence, AI Institute, Shanghai Jiao Tong University, Shanghai, China\\
% {\tt\small \{zhangjinglei168, ltc19940819, kchen, zhang_rui\}@sjtu.edu.cn}

\renewcommand{\shortauthors}{Jinglei Zhang and Tiancheng Lin, et al.}

%%
%% The abstract is a short summary of the work to be presented in the
%% article.
\begin{abstract}
Context-aware methods achieved great success in supervised scene text recognition via incorporating semantic priors from words. We argue that such prior contextual information can be interpreted as the relations of textual primitives due to the heterogeneous text and background, which can provide effective self-supervised labels for representation learning. However, textual relations are restricted to the finite size of dataset due to lexical dependencies, which causes the problem of over-fitting and compromises representation robustness. To this end, we propose to enrich the textual relations via rearrangement, hierarchy and interaction, and design a unified framework called RCLSTR: \textbf{R}elational \textbf{C}ontrastive \textbf{L}earning for \textbf{S}cene \textbf{T}ext \textbf{R}ecognition. Based on causality, we theoretically explain that three modules suppress the bias caused by the contextual prior and thus guarantee representation robustness. Experiments on representation quality show that our method outperforms state-of-the-art self-supervised STR methods. Code is available at \url{https://github.com/ThunderVVV/RCLSTR}.
\end{abstract}

%%
%% The code below is generated by the tool at http://dl.acm.org/ccs.cfm.
%% Please copy and paste the code instead of the example below.
%%
% \begin{CCSXML}
% <ccs2012>
%    <concept>
%        <concept_id>10010147.10010178.10010224.10010240.10010241</concept_id>
%        <concept_desc>Computing methodologies~Image representations</concept_desc>
%        <concept_significance>300</concept_significance>
%        </concept>
%  </ccs2012>
% \end{CCSXML}

% \ccsdesc[300]{Computing methodologies~Image representations}

%%
%% Keywords. The author(s) should pick words that accurately describe
%% the work being presented. Separate the keywords with commas.
% \keywords{contrastive learning, text recognition, relation}
%% A "teaser" image appears between the author and affiliation
%% information and the body of the document, and typically spans the
%% page.
% \begin{teaserfigure}
%   \includegraphics[width=\textwidth]{sampleteaser}
%   \caption{Seattle Mariners at Spring Training, 2010.}
%   \Description{Enjoying the baseball game from the third-base
%   seats. Ichiro Suzuki preparing to bat.}
%   \label{fig:teaser}
% \end{teaserfigure}

% \received{20 February 2007}
% \received[revised]{12 March 2009}
% \received[accepted]{5 June 2009}

%%
%% This command processes the author and affiliation and title
%% information and builds the first part of the formatted document.
\maketitle

\section{Introduction}
\label{sec:intro}

% Contrastive learning methods \cite{he2020momentum,chen2020simple,chen2020improved,caron2020unsupervised} have shown promising improvements in computer vision tasks for natural images. These self-supervised learning methods can achieve good performance using unlabeled data, which is suitable for scene text recognition (STR).  Scene text recognition is a computer vision task that decodes text from images containing text, and usually needs a mass of data for training.  Scene text is very different from natural images and has complex backgrounds. Directly appling contrastive learning method of natural images to scene text images does not consider the characteristics of text, which will bring performance loss. Therefore, we need a contrastive learning methods designed for text images.

Self-supervised learning (SSL), especially contrastive learning methods~\cite{npid,cpc,he2020momentum,chen2020improved,chen2020simple,swav,byol,simsiam}, has achieved great success in computer vision tasks for natural images.
An excellent visual representation learned from unlabeled data is attractive for scene text recognition (STR). Otherwise, a mass of labeled data is usually needed for training to decode the contained text from images~\cite{str1,str2,str3}.  
Directly transferring the contrastive learning methods of natural images to scene text images is sub-optimal since the characteristics of scene text images are quite different from natural images. We argue that text images mainly have the following essential characteristics. First, foreground ($i.e.$, text) and background are heterogeneous in text images, and text recognition relies primarily on text rather than the background. Second, text images are known to have a left-to-right structure. Third, besides the whole image, text images contain the sequence of characters and structure of multi-granularity. Significantly these text characteristics should be fully explored and accordingly propose a new framework of SSL on scene text images.

%Context-aware methods~\cite{context1,context2,fang2021read} achieved great success in supervised scene text recognition via incorporating semantic priors from words. We argue that such contextual information can be interpreted as the relations of textual primitives. 
\begin{figure}[t]
  \centering
  \includegraphics[width=0.9\columnwidth]{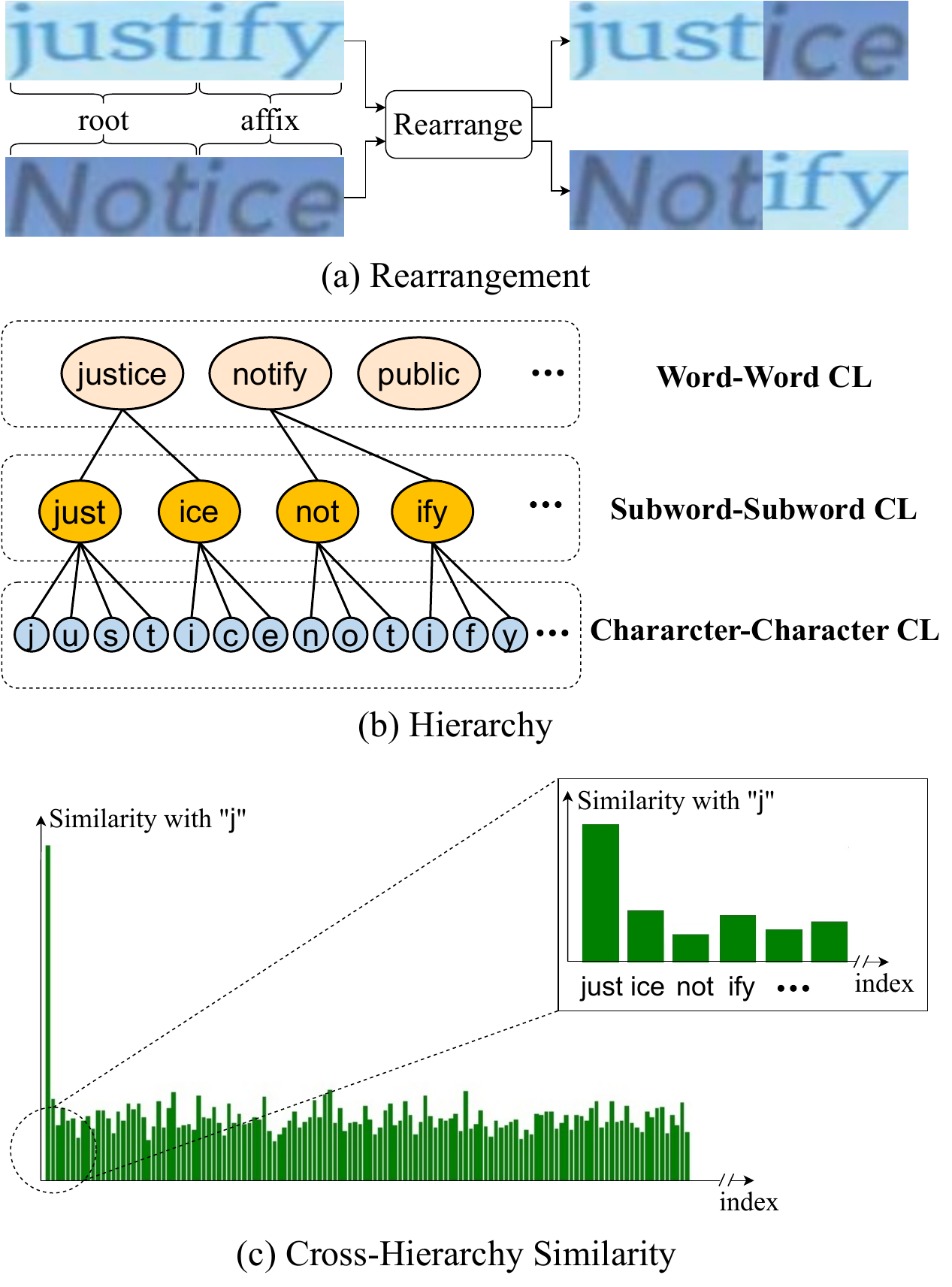}
  \caption{{\bf Textual Relations.} (a) In the dataset, the context is limited. By rearrangement (shuffling and concatenating), we can create a richer context for regularization. (b) Text images naturally have hierarchical features on multiple levels. The most granular level is character. Multiple characters form a subword, and multiple subwords form a word. We use ``CL" to denote contrastive learning. (c) For cross-hierarchy relations, the character presents higher similarity with the subword from the same region than the subwords in other regions. Similarly, the subword shows higher similarity with the word from the same image than the words in other images.}
  \label{fig:motivation}
% \vspace{-0.7cm}
\end{figure}

Some pioneering works~\cite{aberdam2021sequence,liu2022perceiving,yang2022reading} have explored how to construct variants of contrastive learning for text recognition. SeqCLR~\cite{aberdam2021sequence} considers scene text images as a sequence of subwords and thus proposes an instance-mapping function, which makes the atoms of contrastive learning to be sequential frames ($i.e.$, the subwords) rather than images ($i.e.$, the whole image). PerSec~\cite{liu2022perceiving} conducts contrastive learning on low-level and high-level features, aiming to simultaneously learn the representations from stroke and semantic context. DiG~\cite{yang2022reading} directly integrates contrastive learning and masked image modeling (MIM) into a unified model for text recognition, while the gains mainly come from the powerful ViT~\cite{dosovitskiy2020image} and MIM~\cite{xie2022simmim}.
These above self-supervised methods are mainly transferred from natural images and only partially explore the text characteristics. 
Unlike them, our idea is rooted in the supervised text recognition in our community, aiming at fully exploring the characteristics of the text. 
In particular, context-aware methods~\cite{context1,context2,fang2021read} achieved great success by incorporating semantic priors from words in a supervised fashion. 
We argue that such contextual information can be interpreted as the relations of textual primitives, thus, can be utilized in an unsupervised way. 
Unfortunately, textual relations are restricted to the finite size of the dataset, which usually causes the problem of over-fitting due to lexical dependencies~\cite{wan2020vocabulary}. To address this problem, we propose to enrich the textual relations via rearrangement, hierarchy and interaction, resulting in a more complete contrastive mechanism. For ``rearrangement'', text images can be divided and rearranged into new context relations. For ``hierarchy'', there are multi-level relations in text images, such as words, subwords and characters. For ``interaction'', we can leverage the interactions among the objects of different levels, $e.g.$, character-subword and subword-word similarities. 
% Thus, we propose .

%Permutation of different roots and affixes would produce new context relationships.
%In fact, roots and affixes contained in words can be used as primitives, where semantic relationships could be found among these two kinds of primitives.  Besides, roots and affixes have hierarchical relationships with words.

Correspondingly, we propose RCLSTR: Relational Contrastive Learning for Scene Text Recognition, with three novel modules to fully explore the relations in texts. 
% We argue that text images have particular characteristics, such as left-to-right and multi-level structure, which can be used to optimize contrastive learning. 
First, we design a relational regularization module to generate new word images, enriching the variety and diversity of relations. 
% Given a finite training set, the relations are limitedly represented using the existing vocabulary, thus the model might show a poor generalization ability in a new downstream task under changed contextual relations. 
Instead of enumerating all possible relations in texts, which is impractical, we turn to creating new images on-the-fly. The rearrangement of images creates richer context relations.
For example, as shown in Figure~\ref{fig:motivation} (a), words ($e.g.$, ``justify" and ``notice") can be broken up into subwords of roots and affixes and
new words of ``justice" and  ``notify'' can be achieved by rearrangement. 
In practice, we generate new permuted images by horizontal division and concatenation, since text images usually have left-to-right direction.
Note that the position labels of roots and affixes are unavailable in SSL, so the ideal image division is not attainable. Our experiments further study multiple strategies for image division and find that ideal division is not necessary for this module.
% Thus we use KL divergence to constrain the relative relation as regularization, avoiding over-fitting the meaningless words.
% Then the subwords from different words ($i.e.,$ ``just" and ``ice") can also be put together to form a new one  --- ``justice". 
% Such data augmentation is 
% Note that new context relationships are produced at the cost of meaningless words.  Thus, we 
% Therefore, the created new words can enhance the variety and diversity of the relations.
% , which can be regarded as a form of relational regularization.
Second, a hierarchical structure is proposed to conduct representation learning at multiple levels of primitives, which is motivated by the fact that texts have multiple objects with different granularities. 
As shown in Figure~\ref{fig:motivation} (b), at the highest level, one image is taken as a whole to learn the representations of words. The words can be divided into subwords ($e.g.$, roots and affixes) in the middle level, and they work as functional language units from the linguistic perspective~\cite{sennrich2015neural}.
The lowest level of characters is the atomic elements of texts. 
% The middle level includes subwords consisting of several characters, which could be roots or affixes, and they work as useful language units from the linguistic perspective~\cite{sennrich2015neural}.  
We hypothesize that mining multi-level relations in a hierarchical structure could enrich semantic information and enhance representation learning. 
Third, besides the intra-hierarchical relations, we further propose consistency constraints to explore the inter-hierarchical relations. As shown in Figure~\ref{fig:motivation} (c), the characters (at the lowest level) and subwords (at the middle level) from the same locations (in the same images) share similar attributes in color and stroke, thus showing higher similarity in the feature space. The same phenomenon is also found across the levels of subwords and words. Therefore, we are motivated to explicitly constrain the consistency of semantic similarity across the hierarchical levels of text images. We hypothesize that enabling interaction across multiple levels can facilitate the learning of high-quality representations in a more effective manner.

We summarize the contributions of this work as follows:

\begin{itemize}
    \item We propose to explore the relations in text images for self-supervised learning. Text images encode rich contextual information in the relations among textual primitives, which are essential for contrastive learning. 
    \item We propose a novel framework RCLSTR: \textbf{R}elational \textbf{C}ontrastive \textbf{L}earning for \textbf{S}cene \textbf{T}ext \textbf{R}ecognition, which includes three novel modules for exploring relational regularization, hierarchical relations and inter-hierarchy relational consistency.
    \item Our RCLSTR achieves superior performance over the state-of-the-art self-supervised STR methods on representation quality. Moreover, the effectiveness of key model components is verified by the ablation study.
\end{itemize}

\begin{figure*}[t]
\normalsize
  \centering
  \includegraphics[width=\textwidth]{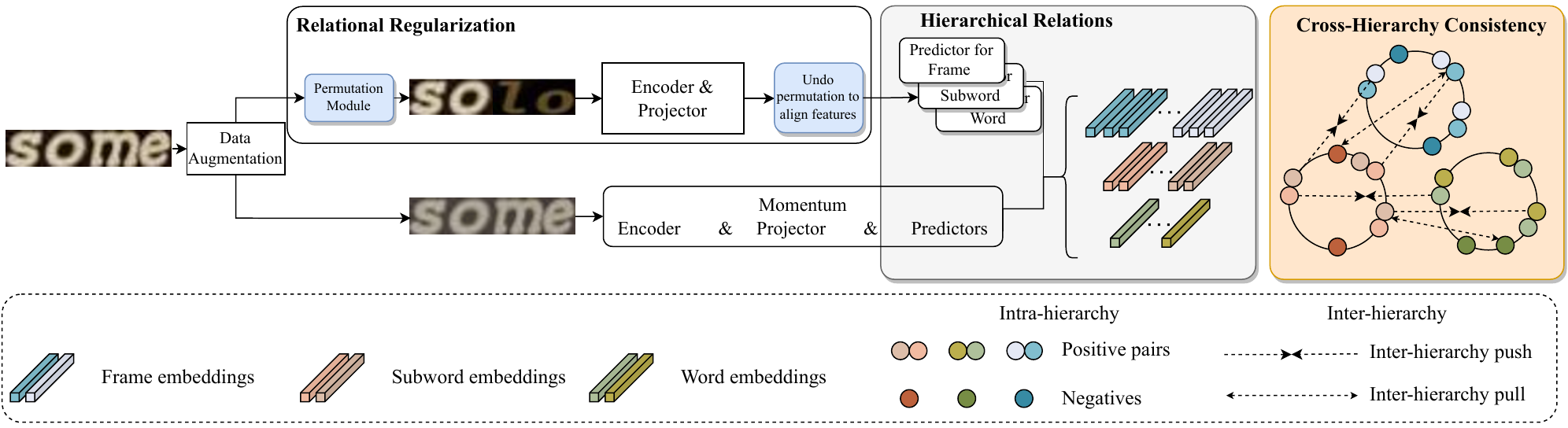}
  \vspace{-0.5cm}
  \caption{\textbf{Block diagram.} Each image in a batch is augmented twice and then fed separately into the online branch (top) and the momentum branch (bottom) of the encoder and projector to create pairs of representation maps. In relational regularization module, we randomly permute the image patches and undo permutation on their features. Next, for the hierarchical contrastive learning of these representations, we apply three predictors that transform them into frames, subwords and words, respectively. In relational consistency module, the corresponding positions on three circles represent the same spatial positions across different hierarchical levels. And we take the corresponding frame \& subword or subword \& word as positive pairs. The diagram uses the corresponding colors to represent the three levels.}
  
  \label{fig:main_fig_framework}
\end{figure*}

\section{Related Work}

\subsection{Self-Supervised Learning}

For natural image self-supervised learning, contrastive learning methods~\cite{npid,cpc,he2020momentum,chen2020improved,chen2020simple,swav,byol,simsiam} show great success, which performs the instance discrimination task to classify different data-augmentation views from the same image into a class. In NPID~\cite{npid}, the task of instance discrimination is proposed, and noise contrastive estimation (NCE) is used for contrastive learning, which is further replaced by InfoNCE~\cite{cpc}. MoCo~\cite{he2020momentum} and SimCLR~\cite{chen2020simple} improve the quality of learned representations, which proposes the momentum encoder and uses a single network with a large batch size, respectively. SwAV~\cite{swav} constrains the consistency of cluster allocation of different data-augmentation views. Recently, BYOL~\cite{byol} and Simsiam~\cite{simsiam} further propose asymmetric frameworks which do not need negative samples. 
% Besides instance-discrimination contrastive learning
More recently, some works~\cite{patacchiola2020self,zheng2021ressl,mitrovic2020representation,wei2020co2} propose to use KL divergence to constrain relative consistency in the form of similarity distribution.
% obtained by two data augmentations to be consistent.
However, these self-supervised methods are designed for natural images, which is quite different from text images. Considering the characteristics of text, we need to specially design the self-supervised method for text images.

%-------------------------------------------------------------------------
\subsection{Self-Supervised Text Recognition}

Some pioneering works~\cite{whatif,aberdam2021sequence,liu2022perceiving,yang2022reading,siman} explored self-supervised methods in text recognition and have achieved promising results. We summarize these methods into three main categories. The first approach is based on contrastive learning. SeqCLR~\cite{aberdam2021sequence} maps sequence features of words to instances as atomic elements of contrast learning. They only consider the text sequence structure. PerSec~\cite{liu2022perceiving} proposed to conduct contrastive learning on the low-level stroke and the high-level semantic features of text images corresponding to visual and semantic information. The second is based on mask image model. DiG~\cite{yang2022reading} proposed a self-supervised framework for text
recognition that combines contrastive learning and masked image models(MIM). Concurrent work~\cite{maskocr} also uses MIM for STR. However, these MIM methods are directly transferred from natural image methods. The third is based on generative learning, and a representative work is  SimAN~\cite{siman}, which proposes to reconstruct the images from the decoupled content and style information.
In sum, the above methods have not fully explored the characteristics of text images. Our approach takes into account the heterogeneity of texts, the left-to-right structure and the hierarchical structure of the sequence to fully explore the text characteristics. We propose a novel contrastive learning framework to enrich the textual relations via rearrangement, hierarchy and interaction.
%MaskOCR
\section{Method}

Based on the structure of MoCo~\cite{he2020momentum}, an efficient and effective baseline, we propose the relational contrastive learning framework for text recognition (RCLSTR).
% which is depicted in Figure ~\ref{fig:main_fig_framework}. 
% We choose MoCo for its representativeness and  
As shown in Figure~\ref{fig:main_fig_framework}, we introduce a novel permutation stage in the online branch (upper branch) to yield horizontal permuted images from the original, which is denoted as relational regularization module (Sect.~\ref{sec:regularization}). In addition, we design a hierarchical structure to learn relations at each level, which is called hierarchical relation module (Sect.~\ref{sec:hierarchical}). Meanwhile, we propose a cross-hierarchy relational consistency module (Sect.~\ref{sec:consistency}) so that the network learns the relation between hierarchies.

% As , we suggest a framework consisting of the multiple blocks. And we detailedly introduce relational regularization, hierarchical relations and relational consistency, and .

%-------------------------------------------------------------------------

\subsection{Preliminaries}

\noindent\textbf{Text recognition framework.} 
As we focus on general contrastive learning for text images, we follow SeqCLR~\cite{aberdam2021sequence} and use a general text recognition framework in ~\cite{baek2019wrong}. This framework is the foundation of many text recognizers, which consists of an encoder and a decoder. In the encoder, we use a Thin Plate Spline (TPS) transformation~\cite{shi2016robust} and a feature extraction network. The decoder can be a CTC-based decoder~\cite{graves2006connectionist} or attention-based decoder~\cite{cheng2017focusing}.
Note that there are other kinds of text recognition architectures~\cite{str3,str2,fang2021read,xie2022toward} in recent research, and it is expected our RCLSTR can also be applied to them.

\noindent\textbf{Contrastive learning.} Contrastive learning methods~\cite{he2020momentum,chen2020simple,chen2020improved,caron2020unsupervised} perform an instance discrimination pretext task in the pre-training phase. This pretext task trains the model to discriminate the positive view from the negative views. The query view $\mathrm{X}_i^q$ and positive view $\mathrm{X}_i^p$ are encoded as $\mathbf{q}$ and $\mathbf{p}$. To avoid the need for large batchsize, we follow MoCo to maintain a queue of size $K$, and there are $K$ negative features $\{ \mathbf{n}_k\}_{k=1}^K$ from other images. Then, the contrastive loss of InfoNCE is written as:
\begin{equation}
\resizebox{0.9\hsize}{!}{$\mathcal{L}_{info}(\mathbf{q}, \mathbf{p}, \mathbf{n}) = -\log \frac{\exp (\mathbf{q}\cdot \mathbf{p} / \tau_{info})}{\sum_{\mathbf{u} \in \{ \mathbf{n}_k\}_{k=1}^K \cup \{\mathbf{p} \}} \exp(\mathbf{q}\cdot \mathbf{u} / \tau_{info})}\,, $}
\label{eq:infoloss}
\end{equation}
where $\tau_{info}$ is a temperature hyper-parameter. This loss function aims to pull closer together features of positive pairs and to push all the other negative examples farther apart.

\noindent\textbf{Naive relational contrastive learning.} 
% We expect the scene text model could learn the relation between two arbitrary views during pre-training.
Relational contrastive learning aims at learning not only the relation between query views and positive views, but also the relation between query views and negative views. Inspired by~\cite{zheng2021ressl,wang2022contrastive,wei2020co2}, we calculate the similarity between the positive and the negatives ($i.e.$ $P$) and that between the query and the negatives ($i.e.$ $Q$). We encourage the agreement of two similarity distributions. Formally, we use symmetric Kullback-Leibler (KL) Divergence as the measure of disagreement, imposing consistency between $P$ and $Q$:
\begin{align}
\begin{split}
Q_i(\mathbf{q}, \mathbf{n}) & = \frac{\exp(\mathbf{q} \cdot \mathbf{n}_i / \tau_{kl})}{\sum_{k=1}^K \exp(\mathbf{q} \cdot \mathbf{n}_k / \tau_{kl})}, \\
P_i(\mathbf{p}, \mathbf{n}) & = \frac{\exp(\mathbf{p} \cdot \mathbf{n}_i / \tau_{kl})}{\sum_{k=1}^K \exp(\mathbf{p} \cdot \mathbf{n}_k / \tau_{kl})}, \\
\mathcal{L}_{kl}(\mathbf{q}, \mathbf{p}, \mathbf{n}) & = \frac{1}{2} D_{\mathrm{KL}} (P \Vert Q) + \frac{1}{2} D_{\mathrm{KL}} (Q \Vert P),
\end{split}
\label{eq:klloss}
\end{align}
where $\tau_{kl}$ is also a temperature hyper-parameter. The total relational loss is a weighted average of the InfoNCE loss term and the KL loss term: 
\begin{equation}
\mathcal{L}_{re}(\mathbf{q}, \mathbf{p}, \mathbf{n}) = \mathcal{L}_{info}(\mathbf{q}, \mathbf{p}, \mathbf{n}) + \alpha \mathcal{L}_{kl}(\mathbf{q}, \mathbf{p}, \mathbf{n}), \label{eq:relationloss}
\end{equation}
where $\alpha$ denotes the coefficient to balance the two terms. The first term is the absolute similarity constraint between $\textbf{q}$ and $\textbf{p}$. The second term is the relative similarity constraint, which aims to keep the similarity distribution consistency of $\textbf{q}$ and $\textbf{p}$ with the negatives. 

However, due to the finite size of the dataset, the textual relations are restricted, and the performance of naive relational contrastive learning is limited. Therefore, we propose relational regularization, hierarchical relation and cross-hierarchy relation modules to learn richer textual relations, building a more complete relational contrastive learning framework.

% We expect the scene text model could learn the relation between two arbitrary views during pre-training.

\begin{figure}[t]
  \centering
  \includegraphics[width=1\columnwidth]{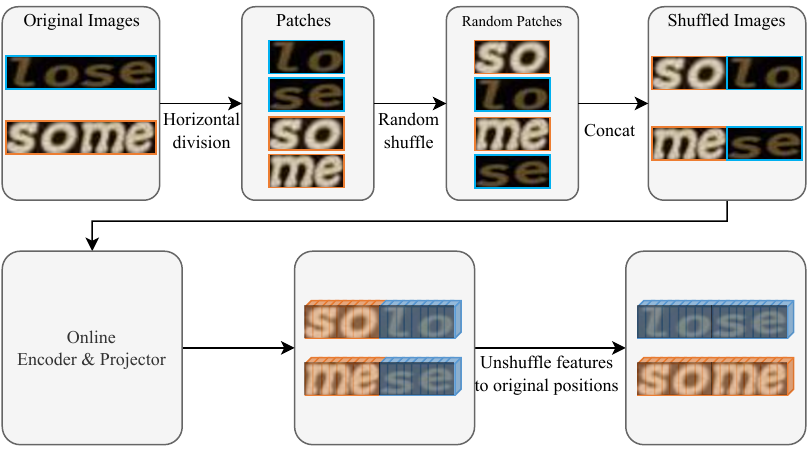}
  % \vspace{-0.7cm}
  \caption{An illustration of the random permutation operation, which generates features for relational regularization.}
  \label{fig:permutation_method}
  % \vspace{-0.7cm}
\end{figure}

\subsection{Relational Regularization}
\label{sec:regularization}

% STR model usually treats each feature in a sequence as the atom for prediction. Previous work~\cite{wan2020vocabulary,zhang2022context} has pointed out that text recognizers are prone to over-dependence on context. It should be noted that the previous methods are used for supervised learning, while our method is proposed for unsupervised learning. Considering the original context is completely dependent on the training set, we propose an effective mechanism to enrich the contextual information. Specifically, we propose to use regularized relational contrastive learning to alleviate this context-dependent problem --- generating the permuted images.
% Generally, the process goes like 1) dividing text images horizontally into several patches, which can be the roots and affixes in words, 2) randomly shuffling and combining patches from different images, and 3) Adding a regularization term into the contrastive learning.
%%REFINE
STR model usually treats each feature in a sequence as the atom for prediction. Previous works~\cite{wan2020vocabulary,zhang2022context} have pointed out that text recognizers are prone to over-dependence on context. It should be noted that the previous methods are used for supervised learning, while our method is proposed for unsupervised learning. In order to alleviate this context-dependent problem, we propose a permutation module to generate new text images. The generated images contain more diversity of context relations, encouraging the encoder not to over-fitting finite contexts in the dataset.
Generally, the process goes like 1) dividing text images horizontally into several patches, 2) randomly shuffling and concatenating patches to generate permuted images, and 3) adding a regularization loss term corresponding to these permuted images. 
% Considering this process may generate meaningless words because roots and affixes labels are not available, we use relative relational contrastive loss for regularization, avoiding over-fitting the meaningless words.

% The permutation operation is performed directly on the input image. As shown in \ref{fig:main_fig_framework}, we take $M$ images $\{\mathbf{x}_i\}_{i=1}^{M}$ as a group. Each image $\mathbf{x} \in \mathbb{R}^{H \times W \times C}$ is reshaped into a sequence of $N$ patches $\mathbf{x}_p \in \mathbb{R}^{H \times W/N \times C}$, and a group of images forms $NM$ patches. Then we randomly arrange these patches and compose a new image for every $N$ patches. Therefore, $M$ original images $\{\mathbf{x}_i\}_{i=1}^{M}$ produce $M$ shuffled images $\{\mathbf{x}_i^{reg}\}_{i=1}^{M}$, where the superscript ``reg" indicates these generated images can serve as a regularization for the textual relations. 
% Note that we only feed $\mathbf{x}^{reg}$ into the online encoder and projector to get frame features. We empirically find such an implementation performs better, which shares the same spirit of multi-cropping strategy~\cite{swav}.
% To aligns all features of permuted images, we should unshuffle them (inverting the random shuffle operation) to put the features position back to their original position.  We denote the resulting features of regularization as $\mathbf{q}^{reg}$. 

%%REFINE
Specifically, the permutation operation is performed directly on the input images, as shown in Figure~\ref{fig:permutation_method}. Firstly, we divide each image horizontally into $N$ patches, where the default $N$ is 2. Next, we take $M$ images as a group to randomly shuffle the $NM$ patches in each group, where the default $M$ is 2. Then, every $N$ patches are concatenated horizontally to make new images. Therefore, we produce shuffled images, denoted as $\{\mathbf{x}^{reg}\}$. 

We only feed $\mathbf{x}^{reg}$ into the online encoder and projector to get frame features, and such an implementation empirically performs better, which shares the same spirit of a multi-cropping strategy~\cite{swav}. To align all features of permuted images, we unshuffle them (inverting the random shuffle operation) to put the features back in their original position. We denote the resulting features of regularization as $\mathbf{q}^{reg}$. The relational contrastive loss with regularization of the permuted images can be written as:
\begin{equation}
\mathcal{L}_{reg}(\mathbf{q}, \mathbf{p}, \mathbf{n}) = \mathcal{L}_{re}(\mathbf{q}, \mathbf{p}, \mathbf{n})  + \mathcal{L}_{re}(\mathbf{q}^{reg}, \mathbf{p}, \mathbf{n}),\label{eq:regularization}
\end{equation}
where $\mathcal{L}_{re}$ is from Equation~\ref{eq:relationloss}. $\mathcal{L}_{reg}$ constrains the invariance of the relation under random permutations. 
The regularization comes from the fact that this constraint on both original and permuted images can force the model not to over-fit the existing contexts.
% and 2) the loss of KL divergence also prevents the model from over-fitting meaningless words.

\begin{figure}[t]
  \centering
  \includegraphics[width=0.8\columnwidth]{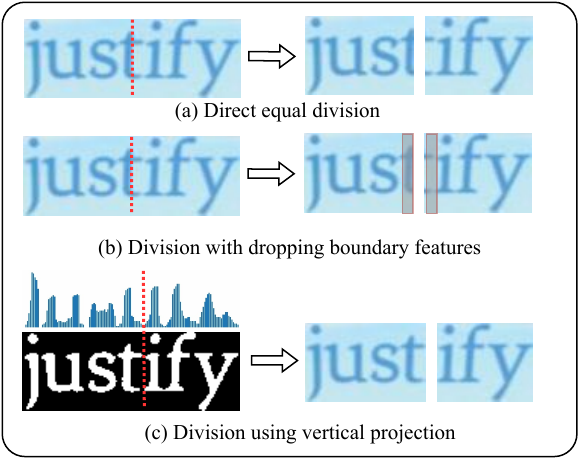}
  \vspace{-0.3cm}
  \caption{Multiple image division strategies. }
  \label{fig:imgdiv2}
  % \vspace{-0.7cm}
\end{figure}

%%REFINE
For the step of horizontal division, since no character position information is available for SSL, we choose equal division as our default setting, which may generate partial characters. And we further study multiple image division strategies (illustrated in Figure~\ref{fig:imgdiv2}). 1) Default direct cutting for equal division. 2) Cutting and dropping boundary features. Since the boundary features may correspond to the characters that are cut, we drop these features when calculating the contrastive loss. 3) Using vertical projection to cut. The vertical projection method can cut from the character gap to avoid cutting the character itself.

% equal division with extra boundary areas. For the second strategy, we preserve areas of extra width at the image dividing boundary and discard the extra redundant features after encoder, which makes the reserved features have full character perception. 

\subsection{Hierarchical Relation}
\label{sec:hierarchical}

Since text images are encoded as sequence features, contrastive learning is applied to the individual elements of the sequence. Considering text words have different granularities in the horizontal direction, we propose a novel hierarchical structure, which maps the features to three levels of frame, subword and word. Thus we conduct hierarchical relational contrastive learning to learn about relations at each level. 

To this end, we use three mapping functions to map the feature sequence into three levels, where representations of different objects are encoded. Specifically, the most fine-grained level is called the frame, which usually contains stroke information of only a portion of the letter. We use the identity function as the frame mapping function. The middle level is called the subword, and it usually contains one or more letters, like roots and affixes. We use an avgpooling layer as the subword mapping function, which map features to $T$ subwords with $T=4$. The highest level is called word, i.e., contains a whole word. And an average function is used as the word mapping function. At each level, we maintain a separate queue of negative features, respectively. We calculate the relational contrastive losses at each level and sum them up:
\begin{equation}
\mathcal{L}_{hier} =  \sum_{h \in H}\mathcal{L}_{reg}(\mathbf{q}^h, \mathbf{p}^h, \mathbf{n}^h),
\end{equation}
where $H=\{frame, subword, word\}$. With the proposed hierarchical relational contrastive learning, the model can learn the frame-frame, subword-subword and word-word relations simultaneously. At each level, we also perform the regularization as in the Equation~\ref{eq:regularization}. 

\subsection{Cross-Hierarchy Relational Consistency}
\label{sec:consistency}

In the previous section, we obtained the features of multiple levels and performed relational contrastive learning within each level. However, there are semantic relations between features across different levels, which are unexplored. Therefore, we propose consistency constraints to learn the relation between neighboring levels. Our default implementation performs frame-subword and subword-word consistency constraints, and we provide the results of other settings in the experiments. As shown in Figure~\ref{fig:main_fig_framework}, for frame-subword relations, since the frame and subword features from the same spatial locations (in the same images) show higher similarity in the feature space, we treat them as positives and treat features in other locations as negatives. And the subword-word positives and negatives are determined in the same way.
For the frame-subword and subword-word relation, we impose the consistency between the similarity distributions by KL loss as the measure of disagreement:
\begin{align}
\begin{split}
\mathcal{L}_{f2s} & = \mathcal{L}_{kl}(\mathbf{q}^{f}, \mathbf{p}^{s}, \mathbf{n}^{s}), \\
\mathcal{L}_{s2w} & = \mathcal{L}_{kl}(\mathbf{q}^{s}, \mathbf{p}^{w}, \mathbf{n}^{w}),
\end{split}
\end{align}
where superscripts of $\{f$, $s$, $w\}$ denote $\{frame$, $subword$, $word\}$, respectively. This loss constrains the relation between each feature and its neighboring upper-level feature. Here we take the positives and negatives from the upper level because we consider that upper-level features contain more semantic information than lower-level.
Finally, the total loss of our relational contrastive learning is formulated as:
\begin{equation}
\resizebox{0.9\hsize}{!}{$\mathcal{L}_{total} = \underbrace{\sum_{h \in H}\mathcal{L}_{reg}(\mathbf{q}^h, \mathbf{p}^h, \mathbf{n}^h)}_{\text{Regularized hierarchical relation}} + \underbrace{\vphantom{\sum_{h \in H}L_{reg}(\mathbf{q}^h, \mathbf{p}^h, \mathbf{n}^h)}\mathcal{L}_{f2s} + \mathcal{L}_{s2w}}_{\text{Cross-hierarchy consistency}}$}
\end{equation}
% \vspace{-0.6cm}

\begin{table*}[t]
\caption{\textbf{Representation quality.} Accuracy(\%) is used to evaluate the quality of representation from encoder, and we train a decoder with labeled data on top of frozen encoder which was pretrained on unlabeled images. Our method with different modules added is compared with the previous methods, where "reg" denotes the relational regularization module, "hier" denotes the hierarchical relation module and "con" denotes the cross-hierarchy relational consistency module. }
% \vspace{-0.5cm}
\begin{center}
    \centering
    \begin{tabular}{clcccccccl}
    \hline
        \multirow{2}{*}{Decoder} & \multirow{2}{*}{Method} & \multicolumn{8}{c}{Scene-Text Dataset}\\ 
        \cline{3-10}
        & & IIIT5K & IC03 & IC13 & SVT & IC15 & SVTP & CUTE80 & Avg \\ \hline
        \multirow{7}{*}{CTC} & SeqCLR~\cite{aberdam2021sequence} & 35.70 & 43.60 & 43.50 & - & - & - & - & -\\ 
        ~ & PerSec-CNN~\cite{liu2022perceiving} & 37.90 & 45.70 & 46.40 & - & - & - & - & -\\ 
        ~ & SeqMoCo w/o KL Loss & 41.63 & 48.21 & 46.50 & 25.35 & 22.05 & 19.53 & 22.22 & 32.21 \\
        ~ & SeqMoCo & 42.97 & 51.44 & 48.37 & 25.35 & 23.01 & 20.62 & 23.26 & 33.57\\ 
        \cline{2-10}
        ~ & Ours based on SeqMoCo & ~ & ~ & ~ & ~ & ~ & ~ & ~ & ~\\ 
        ~ & w/ reg & 48.43 & 58.94 & 54.98 & 35.09 & 26.43 & 26.82 & 29.17 & 39.98\\ 
        ~ & w/ reg \& hier & 51.90 & 61.36 & 59.01 & 38.79 & 30.62 & 30.08 & 30.21 & 43.14\\ 
        ~ & w/ reg \& hier \& con & \textbf{54.83} & \textbf{64.82} & \textbf{60.89} & \textbf{41.58} & \textbf{32.60} & \textbf{34.26} & \textbf{32.64} & \textbf{45.95} \\ \hline
        \multirow{7}{*}{Atten} & SeqCLR~\cite{aberdam2021sequence} & 49.20 & 63.90 & 59.30 & - & - & - & - & -\\ 
        ~ & PerSec-CNN~\cite{liu2022perceiving} & 50.70 & 65.70 & 61.10 & - & - & - & - & -\\ 
        ~ & SeqMoCo w/o KL Loss & 50.97 & 58.36 & 55.86 & 35.55 & 29.42 & 28.53 & 30.56 & 41.32 \\
        ~ & SeqMoCo & 51.83 & 59.75 & 59.90 & 37.40 & 31.73 & 28.99 & 32.29 & 43.13\\ 
        \cline{2-10}
        ~ & Ours based on SeqMoCo & ~ & ~ & ~ & ~ & ~ & ~ & ~ & ~\\ 
        ~ & w/ reg & 56.30 & 67.70 & 63.25 & 41.27 & 35.05 & 36.9 & 37.15 & 48.23\\ 
        ~ & w/ reg \& hier & 59.03 & 71.51 & 67.29 & 46.37 & 38.32 & 36.90 & 36.81 & 50.89\\ 
        ~ & w/ reg \& hier \& con & \textbf{61.07} & \textbf{72.90} & \textbf{68.77} & \textbf{50.54} & \textbf{40.30} & \textbf{40.16} & \textbf{39.24} & \textbf{53.28}\\ \hline
    \end{tabular}
\end{center}

% \vspace{-0.4cm}
\label{tab:representation_quality}

\end{table*}

\subsection{Justification}
\begin{figure}[t]
  \centering
  \includegraphics[width=0.6\columnwidth]{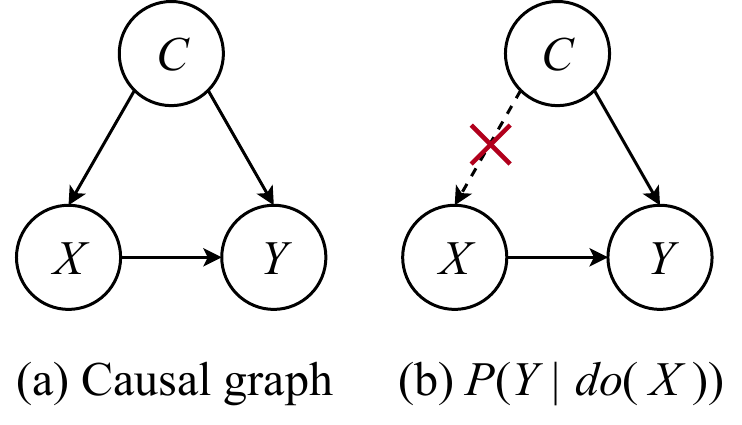}
  % \vspace{-0.4cm}
  \caption{An illustration of causal graph for our framework. }
  \label{fig:causal}
\vspace{-0.5cm}
\end{figure}

As shown in Figure~\ref{fig:causal}, we formulate the SSL framework as a causal graph, which contains three nodes: $X$: scene text images, $Y$: robust representations, and $C$: context information.

\textbf{$X\rightarrow Y$:} This path indicates that the SSL model can learn robust representation for downstream tasks. For example, The success of recent  SeqCLR~\cite{aberdam2021sequence} proves that unsupervised pre-training benefits text representation.

\textbf{$C\rightarrow X$:} This path indicates the generation of scene text images --- combining specific text under some scenes. Some synthetic dataset, like SynthText~\cite{gupta2016synthetic}, is generated in this way.

\textbf{$C\rightarrow Y$:} This path indicates that context information prior in the training dataset can help the learning of robust representations. For example, these context-based STR methods usually utilize that for the prediction of occluded text~\cite{context1,context2,fang2021read}.

In the causal graph, $C$ confounds $X$ and $Y$ via the back-door path $X \rightarrow C \rightarrow Y$, $i.e.,$ learning the representations for STR recognition only based on the dataset prior. The useful context information can be harmful when the data is out-of-distribution, $i.e.$, with a different context prior.
The existence of a back-door path causes a spurious correlation between $X$ and $Y$, which prevents the learning of representations.
An ideal SSL method should capture the true causality between $X$ an $Y$, but the conventional correlation of $P(Y|X)$ fails to do so, as such a spurious correlation is inevitable.
Therefore, we instead seek to use the causal intervention $P(Y|do(X))$, where $do(\cdot)$ is the random controlled trails. As enumerating all textual relations is impossible, we propose the three modules act as the physical intervention, cutting off the confounding effect. In particular, relational regularization module creates new contexts, hierarchical module breaks the context, and cross-hierarchy module learns the correspondence of local-to-global context. They achieve the $do(\cdot)$ operation exactly.

\section{Experiments}

\noindent\textbf{Datasets.} We conduct our experiments on public datasets of scene text recognition. We train on the synthetic dataset SynthText~\cite{gupta2016synthetic}. \textbf{SynthText}~\cite{gupta2016synthetic} is a synthetically generated dataset. It has 5.5M training data once the word boxes are cropped and filtered for non-alphanumeric characters. For evaluation, we use seven real-scene text datasets. \textbf{IIIT5K(IIIT5K-Words)}~\cite{IIIT5K}, \textbf{IC03(ICDAR 2003)}~\cite{ic03}, \textbf{IC13(ICDAR 2013)}~\cite{ic13} and \textbf{SVT(Street View Text)}~\cite{svt} are regular text images, which are nearly horizontal. They contain 3000, 867, 1015 and 647 word images for evaluation, respectively. \textbf{IC15(ICDAR 2015)}~\cite{ic15}, \textbf{SVTP(Street View Text Perspective)}~\cite{svtp} and \textbf{CUTE80}~\cite{cute} are irregular text images. They are mostly perspective text images, and some are blurry or curved. They contain 2077, 645 and 288 word images for evaluation, respectively.

\noindent\textbf{Metrics.} To evaluate performance, we adopt the metrics of word-level accuracy (Acc). Word-level accuracy is the number of correctly predicted words divided by the total number of words. 

\noindent\textbf{Network configurations.} For the network illustrated in Figure~\ref{fig:main_fig_framework}, data augmentation module transforms a given image $\mathrm{X}_i$ in a batch of images into two augmented images $\mathrm{X}_i^a,\mathrm{X}_i^b \in R^{C \times H \times W}$, where we set $H$ as 32, $W$ as 100 and $C$ as 3. We take blocks of transformation (TPS~\cite{shi2016robust}) and feature extraction (ResNet) as the encoder and a two-layer Bidirectional-LSTM (BiLSTM) with 256 hidden units as the projector. For augmented images of both branches, these components extract sequential representations, $\mathrm{R}_i^a, \mathrm{R}_i^b \in R^{F \times T}$, where $F$ is the feature dimension, and $T$ is the number of columns (frames). In our network,  we set $F=256$  and $T=26$ by default. The predictor is a mapping function followed by a fully connected (FC) layer. For the frame level, the mapping function is an identity function. For the subword level, it is an adaptive avgpooling layer that maps frames to $T$ subwords with $T=4$. For the word level, it is an average function. Finally, the feature dimension for contrastive learning is set to 128.

\noindent\textbf{Self-Supervised Pre-Training.} The synthetic dataset SynthText~\cite{gupta2016synthetic} without labels is used for pre-training. We employ an SGD optimizer~\cite{sgd} with a constant learning rate scheduler and train the models for 5 epochs. The training hyperparameters are: the batch size as 32, base learning rate as 1.5e-3, weight decay as 1e-4, momentum for SGD optimizer as 0.9. The pre-training experiments are conducted with 4 GPUs.

\noindent\textbf{Feature Representation Evaluation.} For CTC-based and Attention-based decoders, we inherit the configurations from SeqCLR~\cite{aberdam2021sequence} and PerSec~\cite{liu2022perceiving}. Following the decoder evaluation~\cite{aberdam2021sequence, liu2022perceiving}, during the training for feature representation evaluation, the base encoder is frozen, and we only train a decoder to evaluate the feature representation quality. We employ an Adam optimizer~\cite{adam} and the one-cycle learning rate scheduler~\cite{onecycle} with a maximum learning rate of 5e-4. The training hyperparameters are: the batch size as 256, the number of iterations as 200K, gradient clipping magnitude as 5.

\noindent\textbf{Fine-Tuning Evaluation.} During the training of fine-tuning evaluation, the base encoder is not frozen, and we fine-tune the whole network. Following~\cite{baek2019wrong}, we used ST~\cite{gupta2016synthetic} and MJ~\cite{jaderberg2014synthetic} as the fine-tuning training datasets. An AdaDelta optimizer~\cite{adadelta} and constant learning rate scheduler are employed. The training hyperparameters are: the batch size as 192, the number of iterations as 50K, the base learning rate as 1.0, the decay rate of AdaDelta optimizer as 0.95, gradient clipping magnitude as 5.

\begin{table*}
\caption{\textbf{Fine-tuning results.} Accuracy(\%) of fine-tuning a pretrained model with labeled data. }
% \vspace{-0.5cm}
\begin{center}
    \centering
    \begin{tabular}{clcccccccc}
    \hline
         Decoder & Method & IIIT5K & IC03 & IC13 & SVT & IC15 & SVTP & CUTE80 & Avg\\ \hline
    %     \multirow{5}{*}{CTC} & Supervised baseline & - & - & - & - & - & - & - & - \\ 
    %     ~ & SeqCLR & - & - & - & - & - & - & - & -\\ 
    %     ~ & PerSec-CNN & - & - & - & - & - & - & - & -\\ 
    %     ~ & SeqMoCo & - & - & - & - & - & - & - & -\\ 
    %     ~ & RCLSTR(Ours) & - & - & - & - & - & - & - & - \\ 
    % \hline
        \multirow{5}{*}{Atten} & Supervised baseline & 84.40 & 91.81 & 89.16 & 83.62 & 68.05 & 73.33 & \textbf{71.08} & 80.21 \\ 
        ~ & SeqCLR~\cite{aberdam2021sequence} & 82.90 & 92.20 & 87.90 & - & - & - & - & -\\ 
        ~ & PerSec-CNN~\cite{liu2022perceiving} & 84.20 & - & 88.90 & 82.40 & 68.20 & 73.60 & 68.40 & -\\ 
        ~ & SeqMoCo & 84.40  & \textbf{92.73}  & 89.85  & \textbf{84.54}  & \textbf{69.30}  & 74.88  & 64.81  & 80.08\\ 
        ~ & RCLSTR(Ours) & \textbf{86.03}  & \textbf{92.73}  & \textbf{91.13}  & 83.15  & 69.15  & \textbf{74.88}  & 67.94 & \textbf{80.72} \\ \hline
    \end{tabular}
\end{center}

% \vspace{-0.7cm}
\label{tab:finetune}

\end{table*}

\begin{table*}
\caption{\textbf{Semi-supervised results.} Accuracy(\%) of fine-tuning a pre-training model with 10\% and 1\% of the labeled data.}
% \vspace{-0.5cm}
\begin{center}
    \centering
    \begin{tabular}{clcccccccl}
    \hline
        Fraction & Method & IIIT5K & IC03 & IC13 & SVT & IC15 & SVTP & CUTE80 & Avg\\ \hline
        \multirow{3}{*}{10\%} & Supervised baseline & 70.90  & 83.85  & 79.02  & 66.46  & 49.74  & 50.70  & 47.04 & 63.96 \\ 
        ~ & SeqMoCo & 75.20  & \textbf{87.77}  & 81.87  & 71.41  & 54.48  & 57.98  & 48.78  & 68.21\\ 
        ~ & RCLSTR(Ours) & \textbf{76.80}  & 87.31  & \textbf{82.86}  & \textbf{72.64}  & \textbf{55.31}  & \textbf{60.16}  & \textbf{54.01}  & \textbf{69.87}\\ \hline
        \multirow{3}{*}{1\%} & Supervised baseline & 64.57 & 80.05 & 74.09 & 60.59 & 42.92 & 45.12 & 37.28 & 57.80\\ 
        ~ & SeqMoCo & 65.57  & 81.55  & 74.98  & 62.91  & 48.86  & 53.95  & 37.98 & 60.83 \\ 
        ~ & RCLSTR(Ours) & \textbf{73.73}  & \textbf{86.51}  & \textbf{81.77}  & \textbf{72.80}  & \textbf{51.35}  & \textbf{58.61}  & \textbf{45.99} & \textbf{67.25}\\ \hline
    \end{tabular}

\end{center}
% \vspace{-0.6cm}

% \vspace{-0.4cm}
\label{tab:semi}
\end{table*}

\begin{table*}[!ht]
\caption{The decoder evaluation performance with ViT-based encoder-decoder architecture. Our RCLSTR method with different modules added is compared with the SeqMoCo, where "reg" denotes the relational regularization module, "hier" denotes the hierarchical relation module and "con" denotes the cross-hierarchy relational consistency module.}
    \centering
    \begin{tabular}{cclcccccccc}
    \hline
        \multirow{2}{*}{Encoder-Decoder} & \multirow{2}{*}{Method} & \multirow{2}{*}{Modules} & \multicolumn{8}{c}{Scene-Text Datasets} \\ 
        ~ & ~ & ~ & IIIT5K & IC03 & IC13 & SVT & IC15 & SVTP & CT80 & Avg \\ \hline
        \multirow{4}{*}{\makecell{SATRN$_{small}$\\(ViT-Based)}} & SeqMoCo & - & 62.57 & 75.55 & 70.44 & 63.83 & 44.1 & 46.05 & 35.76 & 56.90  \\ 
        \cline{2-11}
        ~ & \multirow{3}{*}{RCLSTR} & w/ reg & 74.43 & 83.51 & 77.83 & 70.48 & 53.83 & 54.73 & 46.53 & 65.91  \\ 
        ~ & ~ & w/ reg \& hier & \textbf{78.23} & 86.16 & 80.89 & \textbf{75.43} & 58.59 & \textbf{61.24} & 54.51 & 70.72  \\ 
        ~ & ~ & w/ reg \& hier \& con & 78.10 & \textbf{87.31} & \textbf{82.46} & 74.34 & \textbf{59.65} & 60.16 & \textbf{55.21} & \textbf{71.03} \\ \hline
    \end{tabular}

\label{tab:vit}
\end{table*}

\begin{table}[ht]
\caption{The decoder evaluation performance on Chinese and handwritten datasets.}
% \vspace{-0.5cm}
\begin{center}
    \centering
    \begin{tabular}{cccc}
    \hline
         \multirow{2}{*}{Method} & \multirow{2}{*}{Chinese Dataset} & \multicolumn{2}{c}{Handwritten Dataset}\\ 
          ~  & ~ & IAM & CVL \\ \hline 
          SeqMoCo & 47.56 & 56.16 & 77.80  \\
          RCLSTR & \textbf{55.70} & \textbf{62.88} & \textbf{85.92} \\ \hline
           % \\ 
    
    \end{tabular}
\label{tab:handwritten}
\end{center}
% \vspace{-0.7cm}
\end{table}

\subsection{Representation Quality}

For the study of representation quality, the base encoder is unsupervised pre-trained and then frozen. Following SeqCLR~\cite{aberdam2021sequence}, We only train a decoder with labeled data on top of it. We compare our results with other CNN-based SSL methods, and the results are shown in Table~\ref{tab:representation_quality}. Because SeqCLR requires a large batchsize, considering the limitations of hardware, we replace the baseline with MoCo. Based on MoCo~\cite{he2020momentum}, we add the mapping function of SeqCLR~\cite{aberdam2021sequence} and implement the sequential relational contrastive learning method, denoted as the baseline SeqMoCo. Based on SeqMoCo, we add relational regularization module, hierarchical relation module and cross-hierarchy relational consistency module in turn. 
% The performance gain verifies the effectiveness of our three modules. 

The results of representation quality are shown in Table~\ref{tab:representation_quality}. Compared with SeqMoCo without KL loss, SeqMoCo with naive relational contrastive learning achieves limited performance gain. This performance gain is limited by finite dataset relations and suffers from over-fitting due to the lexical dependencies. Compared with SeqMoCo, our method equipped with all three modules further gains an improvement of +12.38\% on average for the CTC-based decoder and +10.15\% for the Attention-based decoder. Also, the effectiveness of three key modules is verified in this table. It should be noted that SeqMoCo is not a stronger baseline (especially in attention-based decoder). Our performance superiority is from the three proposed modules and not from the KL loss or baseline.
% Regularization module gains an improvement of +6.41\% on average for CTC-based decoder and +5.10\% for Attention-based decoder. Hierarchical module gains an improvement of +3.16\% on average for CTC-based decoder and +2.66\% for Attention-based decoder. Hierarchical consistency module gains an improvement of +2.81\% on average for CTC-based decoder and +3.76\% for Attention-based decoder.

\begin{table*}[htb]
\caption{\textbf{Ablations.} (a) Analysis of the setting for hierarchies. Without adding other modules, we try the settings of different combinations for hierarchies. (b) Effect of image division strategies. $^*$: Direct cutting is the default setting of RCLSTR. (c) Effect of consistency constraints. (d) Ablation on hierarchical consistency loss functions.}
\vspace{-0.5cm}
\begin{center}
    \centering
    \begin{tabular}{clcccccccc}
    \hline
         \multirow{3}{*}{(a)} & Hierarchies  & IIIT5K & IC03 & IC13 & SVT & IC15 & SVTP & CUTE80 & Avg\\ \cline{2-10}        
         ~ & w/ subword \& frame  & 50.07 & 62.17 & 58.82 & 36.01 & 30.52 & 27.29 & 28.82 & 41.96\\ 
         ~ & w/ subword \& word   & 56.30& 66.90 & 64.63 & 43.59 & 35.19 & 35.50 & 35.07 & 48.17\\ 
         ~ & w/ subword \& word \& frame & \textbf{58.80} & \textbf{68.51} & \textbf{66.21} & \textbf{47.45} & \textbf{37.65} & \textbf{37.36} & \textbf{39.24} &\textbf{50.75}\\          
    \hline
         \multirow{3}{*}{(b)} & Image Division & ~ & ~ & ~ & ~ & ~ & ~ & ~ & ~ \\ \cline{2-10} 
         ~ & Direct cutting$^*$ & 61.07 & 72.90 & 68.77 & 50.54 & 40.30 & 40.16 & 39.24 & 53.28\\ 
         ~ & Dropping boundary features & 61.33 & 74.51 & 69.56 & 49.00 & 38.85 & 38.14 & 38.54 & 52.85 \\
         ~ & Vertical projection & 60.97 & 72.66 & 68.08 & 53.17 & 40.25 & 40.31 & 39.24 & 53.53 \\
    \hline
        \multirow{3}{*}{(c)} & Cross-Hierarchy Consistency  & ~ & ~ & ~ & ~ & ~ & ~ & ~ & ~\\ \cline{2-10}
         ~ & w/ subword-word  & 59.47 & 70.24 & 66.60 & 47.30 & 36.11 & 38.14 & 32.29 & 50.02\\ 
         ~ & w/ frame-subword  & 59.60 & 71.28 & 67.00 & 50.08 & 38.81 & \textbf{41.24} & 38.19 & 52.31\\ 
         ~ & w/ subword-word \& frame-subword  & \textbf{61.07} & \textbf{72.90} & \textbf{68.77} & \textbf{50.54} & \textbf{40.30} & 40.16 & \textbf{39.24} & \textbf{53.28}\\ 
         % ~ & w/ subword-word \& frame-subword \& frame-word & 61.4 & 74.39 & 68.57 & 50.85 & 39.09 & 40.62 & 39.24 & 53.45 \\
    \hline
         \multirow{3}{*}{(d)} & \multicolumn{3}{l}{Cross-Hierarchy Consistency Loss Function}  & ~ & ~ & ~ & ~ & ~ & ~\\ \cline{2-10}
         ~ & InfoNCE  & 60.60 & \textbf{73.82} & \textbf{70.15} & 48.38 & 39.38 & \textbf{40.16} & 37.15 & 52.81 \\ 
         ~ & KL  & \textbf{61.07} & 72.90 & 68.77 & \textbf{50.54} & \textbf{40.30} & \textbf{40.16} & \textbf{39.24} & \textbf{53.28}  \\ 
         ~ & RE  & 59.93 & 71.05 & 67.39 & 50.08 & 38.42 & 39.69 & 36.46 & 51.86 \\ \hline
    \end{tabular}

\end{center}
\vspace{-0.3cm}

% ``Image Division" represents the different image division strategies in Figure~\ref{fig:imgdiv2}. ``Hierarchies" represents different combinations of hierarchies for contrastive learning. ``Cross-Hierarchy Consistency" represents different combinations of consistency constraints between hierarchies. ``Cross-Hierarchy Consistency Loss Function" represents different kinds of consistency loss functions for cross-hierarchy consistency module. All results are on representation qualities (decoder evaluation accuracy).}
% \vspace{-0.7cm}
\label{tab:ablation_good}
\end{table*}

\subsection{Fine-tuning}

We further unfreeze the parameters of the encoder and fine-tune it with the decoder. Table~\ref{tab:finetune} shows the performance comparison between our RCLSTR and other methods. ``Supervised baseline" does not perform self-supervised pre-training, in which parameters are randomly initialized. Compared with SeqMoCo, our method gains an improvement of average performance. Compared with SeqCLR~\cite{aberdam2021sequence} and PerSec~\cite{liu2022perceiving}, our RCLSTR can outperform them in most datasets. These results demonstrate that the image encoder learned by RCLSTR benefits downstream recognition fine-tuning. 

\subsection{Semi-Supervised Learning}

We further evaluate our method by considering semi-supervised settings. We use the same encoders as before, which were pre-trained on the unlabeled data, and let the whole network be fine-tuned using 1\% or 10\% of the labeled dataset. We use the same randomly selected data for all the experiments.

Table~\ref{tab:semi} compares our method with SeqMoCo and supervised baseline training. As can be seen, RCLSTR achieves better performance on average under different amount of labeled data. Our method succeeds in significantly improving the results of SeqMoCo. Compared with SeqMoCo, our method gains an improvement of +1.66\% on average for 10\% labeled data and +6.42\% for 1\% labeled data. These results verify that the representations learned by RCLSTR benefit the learning from insufficient data.

% vit
% handwriting

\subsection{Results on ViT-based Encoder-Decoder}

We evaluate our RCLSTR method on the ViT-based encoder-decoder architecture, and the results are summarized in Table~\ref{tab:vit}. We choose the small version of SATRN~\cite{satrn} to verify the effectiveness of our method on the ViT encoder. We perform SeqMoCo and RCLSTR pre-training, and freeze encoder to perform decoder evaluation. And we add three modules in turn. The performance gain verifies that our three modules can also be effective for the ViT-based encoder. 

\subsection{Results on More Languages and Types of Text Image Datasets}

The condition of using our method only assumes that the text is horizontal, so it is also useful for other languages ($e.g.$, Chinese) and other fonts ($e.g.$, handwritten). 
For the Chinese document dataset~\cite{chen2021benchmarking}, we perform SeqMoCo and RCLSTR pre-training, and freeze encoder to perform decoder evaluation, the accuracies are summarized in Table~\ref{tab:handwritten}. RCLSTR achieved superior performance on Chinese datasets. Since Chinese Text Images have left-to-right structures and horizontal multi-grained hierarchies, RCLSTR can also facilitate self-supervised learning of their features.

We further evaluate our RCLSTR method on the handwritten datasets, comparing its performance with SeqMoCo. We consider the English handwritten datasets IAM~\cite{iam} and CVL~\cite{cvl}, and the results are summarized in Table~\ref{tab:handwritten}. Compared with SeqMoCo, RCLSTR achieves better performance on these two datasets and gains an improvement of +6.72\% for IAM and +8.12\% for CVL. Although handwritten fonts have a certain irregularity, our RCLSTR can also utilize their horizontal and multi-hierarchical structure to facilitate feature learning.

\begin{figure}[t]
\centering
\includegraphics[width=1\columnwidth]{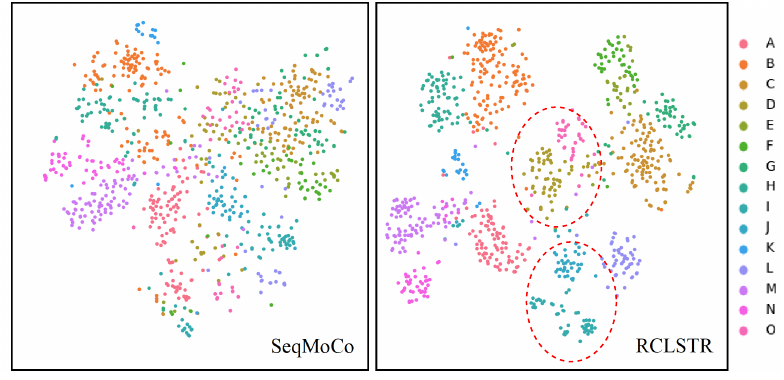}
% \vspace{-0.8cm}
\caption{t-SNE results.}
\label{fig:fig_tsne}
\vspace{-0.7cm}
\end{figure}

\subsection{Visualization}

%%REFINE
In Figure~\ref{fig:fig_tsne}, we use t-SNE~\cite{tsne} to visualize the final features of IIIT5K~\cite{IIIT5K} images corresponding to two methods, $i.e.$, SeqMoCo (baseline) and the proposed RCLSTR, in which features for attention-based decoder are visualized by attaching
character labels to frame features. As can be seen, our method mines intra-class relation to cluster characters of the same class. Besides, our method also mines the inter-class relations, where the clusters with similar-looking characters ($e.g.$, I\&J and D\&O) are close.

\section{Ablation Study}

% \textbf{Effect of modules without KL loss.} Under the condition of removing KL loss, we test the effect of our three modules by only using InfoNCE loss. Under this configuration, we add our three modules in turn. As can be seen in Table~\ref{tab:nokl}, every modules also brings performance gain. These results verify that the benefits of three modules do not depend on the loss type. For relational regularization module, the feature from rearrangement can still provide regularization under InfoNCE loss. Besides, multi-level InfoNCE yields better representations than single-level. And the absolute local-to-global consistency also facilitates hierarchical interaction.
First, we analyze the setting for hierarchies. Without adding other modules, we try the settings of different combinations for hierarchies, and the results are shown in Table~\ref{tab:ablation_good} (a). We can observe that the setting of subword and word gains higher performance than subword and frame. The best performance is achieved by learning at all three levels. In the following ablation, we used all three levels as the default setting to explore variants of other modules.

%%REFINE
\textbf{Effect of image division strategies.} We study how different image division strategies affect the effectiveness of relational regularization. The results under different image division strategies (as shown in Figure~\ref{fig:imgdiv2}) are summarized in Table~\ref{tab:ablation_good} (b). 
Under the condition of no available character positions for SSL, the process of direct division and concatenation creates richer context, but meaningless images (such as partial characters from non-ideal image division) are generated. There is a trade-off between context diversity and non-ideal image division. Since the goal of relational regularization is to avoid over-fitting the original context, the effectiveness mainly comes from more diversity of contexts, and it is insensitive to the non-ideal image division boundaries. We find that dropping the boundary features has a similar performance. And the vertical projection method to avoid cutting character also has no significant performance gap. 
% We can find that division with extra area has similar performance with direct cutting, which shows that non-ideal image division boundaries have little affect on the learning of feature relations. 

\textbf{Effect of consistency constraints.} In Table~\ref{tab:ablation_good} (c), we impose subword-word and frame-subword consistency constraints separately, and we find that frame-subword consistency brings the most performance gain, which indicates that the more granular consistency is more useful for learning text representations. 

\textbf{Ablation on hierarchical consistency loss functions.} By default, we use KL-divergence loss to constrain the consistency between hierarchies. As shown in Table~\ref{tab:ablation_good} (d), we test other consistency loss functions, such as InfoNCE loss in Equation~\ref{eq:infoloss} and relational loss (RE) in Equation~\ref{eq:relationloss}. For cross-hierarchy relations, local and global features do not have absolute consistency and only have relative similarities. Thus KL loss is better than InfoNCE and RE ($i.e.$ InfoNCE + KL) loss.

\vspace{-0.5cm}

\section{Conclusions}
This work proposes a novel framework, Relational Contrastive Learning for Scene Text Recognition (RCLSTR). To take advantage of contextual priors in STR, we argue that contextual information can be interpreted as the relations of textual primitives and utilized in an unsupervised way. In this framework, the relations in text images are fully explored by three modules. The relational regularization module is proposed to enrich the intra- and inter-image context relations. The hierarchical relational module for relational contrastive learning can capture multi-granularity representations. Additionally, the cross-hierarchy relational consistency module is designed for the interactions across different hierarchical levels in scene text images. Experiments on representation quality verify the superiority of our RCLSTR method.

\begin{acks}
This work was supported in part by NSFC 62171282, Shanghai Municipal Science and Technology Major Project (2021SHZDZX0102), 111 project BP0719010, STCSM 22DZ2229005.
\end{acks}

% Our future work is to apply RCLSTR to more text recognition framework with more advanced encoder, such as ViT.
% The main limitation of the current RCLSTR is that the relations is only utilized during pre-training. However, these relations are inherent in the text images and they 
% Note that textual relations can also provide useful constraints in the training of downstream task, which will be our future work.
% Therefore, our future work will mainly focus on enhancing model performance on downstream tasks by utilizing relations in text images.

%%
%% The next two lines define the bibliography style to be used, and
%% the bibliography file.
\bibliographystyle{ACM-Reference-Format}
\bibliography{sample-base}

\clearpage
\appendix

\section{Visualization Results of Regularized Samples}

\begin{figure*}[t]
\normalsize
  \centering
  \includegraphics[width=\textwidth]{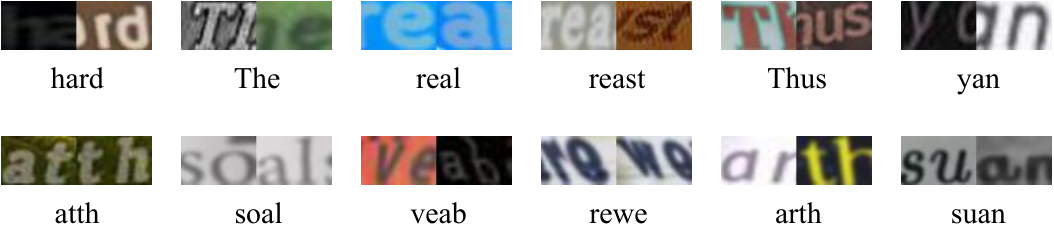}
  \vspace{-0.3cm}

  \caption{\textbf{Visualization results of the regularized samples.} In relational regularization module, we randomly permute the image patches, which generates images with new contexts. }
  \vspace{-0.3cm}
  \label{fig:pm_sample}
\end{figure*}

In Figure~\ref{fig:pm_sample}, we show some samples of randomly permuted images in the relational regularization module. The permutation generates new word images with new contexts. The meaningful generated images are an extension of the contexts. However, non-ideal image divisions (such as partial characters or unaligned boundaries) are generated. As we analyzed in the ablation section, there is a trade-off between context diversity and non-ideal image division. Since relational regularization aims to avoid over-fitting the original context, the effectiveness mainly comes from more diversity of contexts, and our ablation finds that ideal division is not necessary for the relational regularization module.

% \subsection{Analysis of Permutation Strategy}

\begin{table*}[htb]
\caption{\textbf{More results of cross-hierarchy consistency.} ``Cross-Hierarchy Consistency" represents different combinations of consistency constraints between hierarchies. Here we supplement the result with adding frame-word consistency constraint. All results are on representation qualities (decoder evaluation accuracy).}
\vspace{-0.3cm}

\begin{center}
    \centering
    \begin{tabular}{ccccccccccc}
    \hline
         Cross-Hierarchy Consistency & IIIT5K & IC03 & IC13 & SVT & IC15 & SVTP & CUTE80 & Avg\\ \hline
         w/ subword-word \& frame-subword  & 61.07 & 72.90 & 68.77 & 50.54 & 40.30 & 40.16 & 39.24 & 53.28\\
         w/ subword-word \& frame-subword \& frame-word & 61.40 & 74.39 & 68.57 & 50.85 & 39.09 & 40.62 & 39.24 & 53.45 \\
           
    \hline
    \end{tabular}
\label{tab:cross}
\end{center}
% \vspace{-0.5cm}
\vspace{-0.3cm}
\end{table*}

\section{More Results on Cross-Hierarchy Consistency}

In Table~\ref{tab:cross}, we provide the performance of adding frame-word consistency constraints. We find that there is no significant performance improvement with adding frame-word consistency constraint. Considering the semantic similarity between the frame and word levels is relatively distant, the consistency between the two might be not so important.

\section{Memory Bank Size}
In Table~\ref{tab:mem}, we compare the average accuracy(\%) of decoder evaluation under different settings of memory bank size (number of negatives).  As can be seen, the performance is positively correlated to memory bank size, which is similar to MoCo. The default memory size setting of RCLSTR is 65536, and all other experiments are performed with this setting.
% In Table~\ref{tab:mem}, we compare the average accuracy(\%) of decoder evaluation under different settings of memory bank size (number of negatives).  As can be seen, the performance is positively correlated to memory bank size, which is similar to MoCo. The default memory size setting of RCLSTR is 65536, and all other experiments are performed with this setting.

\begin{table}[ht]
\caption{Comparison of different memory bank size.}
\vspace{-0.3cm}
\begin{tabular}{cccc}
    \hline
         Size & 4096 & 16384 & 65536\\ \hline
         Avg Acc & 49.65 & 52.13 & 53.28\\ \hline  
\end{tabular}
\vspace{-0.3cm}
\label{tab:mem}
\end{table}

\section{Computational Cost \& Model Size}
\noindent\textbf{Computational Cost.} In Table~\ref{tab:time}, we compare the per-iteration forwarding time(s) for pre-training SeqMoCo and RCLSTR on the same training machine. The results show that RCLSTR needs only a tiny amount of extra computations. It should be noted that this time may be different on other kinds of hardware, and our purpose is to illustrate the relative computational cost.

\begin{table}[ht]
\caption{Per-iteration forwarding time(s).}
\vspace{-0.3cm}

\begin{tabular}{ccc}
    \hline
         Method & SeqMoCo & RCLSTR\\ \hline
         Time(s) & 0.148 & 0.178 \\ \hline
    
    \end{tabular}
\vspace{-0.3cm}
\label{tab:time}
\end{table}

\noindent\textbf{Model Size.} We build the encoder and decoder following SeqCLR, so the model sizes of SeqCLR and ours are fair. The Persec decoder is the same as ours. Because some parameter nums are unreleased, Persec-CNN has an unknown size for the encoder. It should be noted that our model also exceeds Persec-ViT(as shown in Table~\ref{tab:persec-vit}), which uses a stronger encoder than Persec-CNN. So our performance is better than PerSec. To sum up, RCLSTR is relatively fair in terms of model size, and our performance improvement is not from a larger model size.

\begin{table}[ht]
\caption{Additional comparison with Persec-ViT.}
\vspace{-0.3cm}
\begin{tabular}{clccc}
    \hline
         Decoder & Method & IIIT5K & IC03 & IC13\\ \hline
         \multirow{2}{*}{Atten} & Persec-ViT~\cite{liu2022perceiving} & 52.30 & 66.60 & 62.30 \\ 
         ~ & RCLSTR & \textbf{61.07} & \textbf{72.90} & \textbf{68.77} \\ \hline
    
    \end{tabular}
\label{tab:persec-vit}
\end{table}

\section{More Implementation Details}
\subsection{Text Recognition Scheme}
We use the ``encoder-decoder" text recognition network. In the encoder, we use a transformation and a feature extraction network. The decoder can be a CTC-based decoder or attention-based decoder. 

The transformation stage transforms a cropped image into a normalized image. Because the input image may contain text in a non-axis aligned layout, as often occurs in scene text images, the transformation is necessary. We follow ~\cite{aberdam2021sequence} and utilize the Thin Plate Spine (TPS) transformation~\cite{shi2016robust}, which is a variant of the spatial transformer network~\cite{shi2016robust}. This transformation first uses a CNN of 4 layers to detect a pre-defined number of fiducial points at the top and bottom of the text region. Then, a smooth spline interpolation is applied between the obtained points to map the predicted textual region to a constant pre-defined size.

In the feature extraction stage, we use a ResNet~\cite{resnet} of 29 layers, which is the same as ~\cite{aberdam2021sequence}. For Bidirectional-LSTM (BiLSTM), we follow ~\cite{aberdam2021sequence} to use 2 layers of BiLSTM, and the hidden size is 256. We also follow ~\cite{aberdam2021sequence} to build Connectionist Temporal Classification (CTC) based and attention-based decoder to decode the predictions from the sequential features. 

\subsection{Data Augmentation}
For the data augmentation, we follow SeqCLR~\cite{aberdam2021sequence}, and our augmentation consists of a random subset of the linear contrast, blur, sharpen, crop, perspective transform and piecewise affine. The augmentation procedure is implemented using the imgaug augmentation package, which is used to augment each image twice for self-supervised learning. The pseudo-code for augmentation written with imgaug package is as shown in Algorithm~\ref{alg:aug}.

\subsection{Self-Supervised Pre-Training} 
The goal of self-supervised pre-training is to pre-train the weights of the feature encoder. We use TPS~\cite{shi2016robust} and ResNet~\cite{resnet} of 29 layers as the feature encoder. And we use BiLSTM as a projector, and the hidden size is 256. The projector is followed by mapping functions and FC layers that act as predictors. In the pre-training stage, we use a pre-trained TPS module and freeze its weight. The projector and predictors are auxiliary networks that are
discarded entirely after the pre-training stage. After pre-training, we only use the pre-trained weights of the feature encoder. 

\subsection{Feature Representation Evaluation} 
During the training for feature representation evaluation, the base encoder is frozen, and we only train a decoder. For CTC-based and Attention-based decoders, we inherit the configurations from SeqCLR~\cite{aberdam2021sequence}. The paper has illustrated the training and testing settings.

\subsection{Fine-Tuning Evaluation} 
For the training of fine-tuning evaluation, the base encoder is not frozen, and we fine-tune the whole network. We inherit the configurations of CTC-based and Attention-based decoders from SeqCLR~\cite{aberdam2021sequence}. The training and testing settings are illustrated in the paper.

\subsection{Semi-Supervised Evaluation} 
During the training of semi-supervised evaluation, the base encoder is not frozen, and we fine-tune the whole network. Following ~\cite{baek2019wrong}, we used ST~\cite{gupta2016synthetic} and MJ~\cite{jaderberg2014synthetic} as the fine-tuning training data sets and used 1\% or 10\% of labeled data of them. An AdaDelta optimizer~\cite{adadelta} and constant learning rate scheduler are employed. The training hyperparameters are: the batch size as 192, the number of iterations as 5K, the base learning rate as 1.0, the decay rate of AdaDelta optimizer as 0.95, gradient clipping magnitude as 5.

\begin{algorithm}[ht]
\caption{Pseudocode of data augmentation.}
\label{alg:aug}
\definecolor{codeblue}{rgb}{0.25,0.5,0.5}
\lstset{
  backgroundcolor=\color{white},
  basicstyle=\fontsize{7.2pt}{7.2pt}\ttfamily\selectfont,
  columns=fullflexible,
  breaklines=true,
  captionpos=b,
  commentstyle=\fontsize{7.2pt}{7.2pt}\color{codeblue},
  keywordstyle=\fontsize{7.2pt}{7.2pt},
%  frame=tb,
}
\begin{lstlisting}[language=Python]
from imgaug import augmenters as iaa
iaa.Sequential([iaa.SomeOf((1, 5),
[
  iaa.LinearContrast((0.5, 1.0)),
  iaa.GaussianBlur((0.5, 1.5)),
  iaa.Crop(percent=((0, 0.4),
                   (0, 0),
                   (0, 0.4),
                   (0, 0.0)),
                   keep_size=True),
  iaa.Crop(percent=((0, 0.0),
                  (0, 0.02),
                  (0, 0),
                  (0, 0.02)),
                  keep_size=True),
  iaa.Sharpen(alpha=(0.0, 0.5),
            lightness=(0.0, 0.5)),
  iaa.PiecewiseAffine(scale=(0.02, 0.03),
                    mode='edge'),
  iaa.PerspectiveTransform(
                    scale=(0.01, 0.02)),
],
random_order=True)])
\end{lstlisting}
\end{algorithm}

\subsection{Algorithm Pseudocode}

Algorithm~\ref{alg:code} provides the pseudo-code of RCLSTR for the pre-training task. As shown in the pseudo-code, we use random permutation for relational regularization. The regularized relational contrastive losses at the frame, subword and word level are calculated separately. Besides, the cross-hierarchy consistency losses for frame-to-subword and subword-to-word are proposed. The final loss is the sum of the regularized relational loss at each level and the relational consistency losses across levels.

\subsection{Evaluation Variance}

We find that the evaluation results may have some tiny variance in the experiment. The evaluation variance may be due to incomplete evaluation protocols. We look forward to future community work to complete the evaluation protocol of this field.

\begin{algorithm*}[p]
\caption{Pseudocode of RCLSTR in a PyTorch-like style.}
\label{alg:code}
\definecolor{codeblue}{rgb}{0.25,0.5,0.5}
\lstset{
  backgroundcolor=\color{white}, 
  basicstyle=\fontsize{8pt}{8pt}\ttfamily\selectfont,
  columns=fullflexible,
  breaklines=true,
  captionpos=b,
  commentstyle=\fontsize{7.2pt}{7.2pt}\color{codeblue},
  keywordstyle=\fontsize{7.2pt}{7.2pt},
%  frame=tb,
}
\begin{lstlisting}[language=python]
# f_q, f_k: online and momentum networks for query and key
# queue: dictionary as a queue of K keys (CxK)
# m: momentum; t: temperature

f_k.params = f_q.params  # initialize
for x in loader:  
    x_q = aug(x)  # a randomly augmented version
    x_k = aug(x)  # another randomly augmented version
    
    # relational regularization of random permutation
    x_q_reg = aug_pm(x_q) 

    frame_q, subword_q, word_q = f_q.forward(x_q)  
    frame_k, subword_k, word_k = f_k.forward(x_k) 

    frame_q_reg, subword_q_reg, word_q_reg = f_q.forward(x_q_reg)  # queries of regularized samples
    
    # calculate relational contrastive loss for each level
    L_frame = relational_loss(frame_q, frame_k, frame_queue)
    L_frame_reg = relational_loss(frame_q_reg, frame_k, frame_queue)
    
    L_subword = relational_loss(subword_q, subword_k, subword_queue)
    L_subword_reg = relational_loss(subword_q_reg, subword_k, subword_queue)
    
    L_word = relational_loss(word_q, word_k, word_queue)
    L_word_reg = relational_loss(word_q_reg, word_k, word_queue)
    
    # calculate cross-hierarchy consistency loss
    L_f2s = relational_loss(frame_q, subword_k, subword_queue)
    L_s2w = relational_loss(subword_q, word_k, word_queue)
    
    # total loss
    loss = L_frame + L_frame_reg + L_subword + L_subword_reg + L_word + L_word_reg + L_f2s + L_s2w
           

    # SGD update: query network
    loss.backward()
    update(f_q.params)

    # momentum update: key network
    f_k.params = m*f_k.params+(1-m)*f_q.params

    # update dictionary, enqueue and dequeue the current minibatch
    enqueue_dequeue(frame_queue, frame_k)  
    enqueue_dequeue(subword_queue, subword_k)
    enqueue_dequeue(word_queue, word_k)

def relational_loss(q, k, queue):
    # positive logits: Nx1
    l_pos = bmm(q.view(N,1,C), k.view(N,C,1))
    # negative logits: NxK
    l_neg = mm(q.view(N,C), queue.view(C,K))

    # logits: Nx(1+K)
    logits = cat([l_pos, l_neg], dim=1)

    # InfoNCE contrastive loss
    labels = zeros(N)  # positives are the 0-th
    InfoNCE_loss = CrossEntropyLoss(logits/t, labels)
    
    # KL contrastive loss
    similarity_q = mm(q.view(N,C), queue.view(C,K))
    similarity_k = mm(k.view(N,C), queue.view(C,K))
    KL_loss = 0.5 * kl(similarity_q, similarity_k) + 0.5 * kl(similarity_k, similarity_q)    
    return InfoNCE_loss + KL_loss
    
\end{lstlisting}
\end{algorithm*}

%%
%% The next two lines define the bibliography style to be used, and
%% the bibliography file.
% \bibliographystyle{ACM-Reference-Format}
% \bibliography{sample-base}

% \end{document}
% \endinput
%%
%% End of file `sample-sigconf.tex'.

\end{document}